\documentclass[]{aastex631}

\usepackage{tcolorbox}
\tcbuselibrary{listings}

\begin{document}

\title{Galactic ChitChat: Using Large Language Models to Converse with Astronomy Literature}

\author{Ioana Ciuc\u{a}}
\affiliation{ \\
Research School of Astronomy \& Astrophysics, Australian National University, \\ 
Cotter Rd., Weston,  ACT 2611, Australia. \\
}

\affiliation{ \\
School of Computing, Australian National University, \\ 
Acton, ACT 2601, Australia. \\
}

\author{Yuan-Sen Ting}
\affiliation{ \\
Research School of Astronomy \& Astrophysics, Australian National University, \\ 
Cotter Rd., Weston, ACT 2611, Australia. \\
}
\affiliation{ \\
School of Computing, Australian National University, \\ 
Acton, ACT 2601, Australia. \\
}



\begin{abstract}
We demonstrate the potential of the state-of-the-art OpenAI GPT-4 large language model to engage in meaningful interactions with Astronomy papers using in-context prompting. To optimize for efficiency, we employ a distillation technique that effectively reduces the size of the original input paper by 50\%, while maintaining the paragraph structure and overall semantic integrity. We then explore the model's responses using a multi-document context (ten distilled documents). Our findings indicate that GPT-4 excels in the multi-document domain, providing detailed answers contextualized within the framework of related research findings. Our results showcase the potential of large language models for the astronomical community, offering a promising avenue for further exploration, particularly the possibility of utilizing the models for hypothesis generation.
\end{abstract}

$$


\section{Introduction}
\label{sec:intro}
Large language models (LLMs) have significantly advanced natural language processing, allowing machines to process and generate intricate text with remarkable quality \citep[e.g.,][]{Devlin2018, brown2020language, Chowdhery2022, bubeck2023sparks}. In this study, we employ `in-context prompting' \footnote{While developing this project, we came across \texttt{paperqa} \citep{white2023}, which performs QA on academic papers, similar to our approach. However, we employ a distillation technique and consider the chat history.} for the SOTA GPT-4 model \citep{OpenAIGPT4}  as a practical solution to tap into the model's capabilities without the need for fine-tuning \citep[e.g.,][]{howard2018, wei2022finetuned} or incorporating specialized adapters \citep[e.g,][]{pfeiffer2020, Hu2021}. This technique enables us to harness the potential of GPT-4, which has not been specifically trained for Astronomy, to engage with Astronomy literature, a field characterized by its complexity and richness of information. Our goal for the future is to use LLMs in conjunction with expert knowledge to identify possible gaps in our knowledge base and generate new scientific hypotheses in Astronomy.
\section{Methods}

\label{methods}

\begin{figure}[ht!]
\centering
 \includegraphics[width=\hsize]{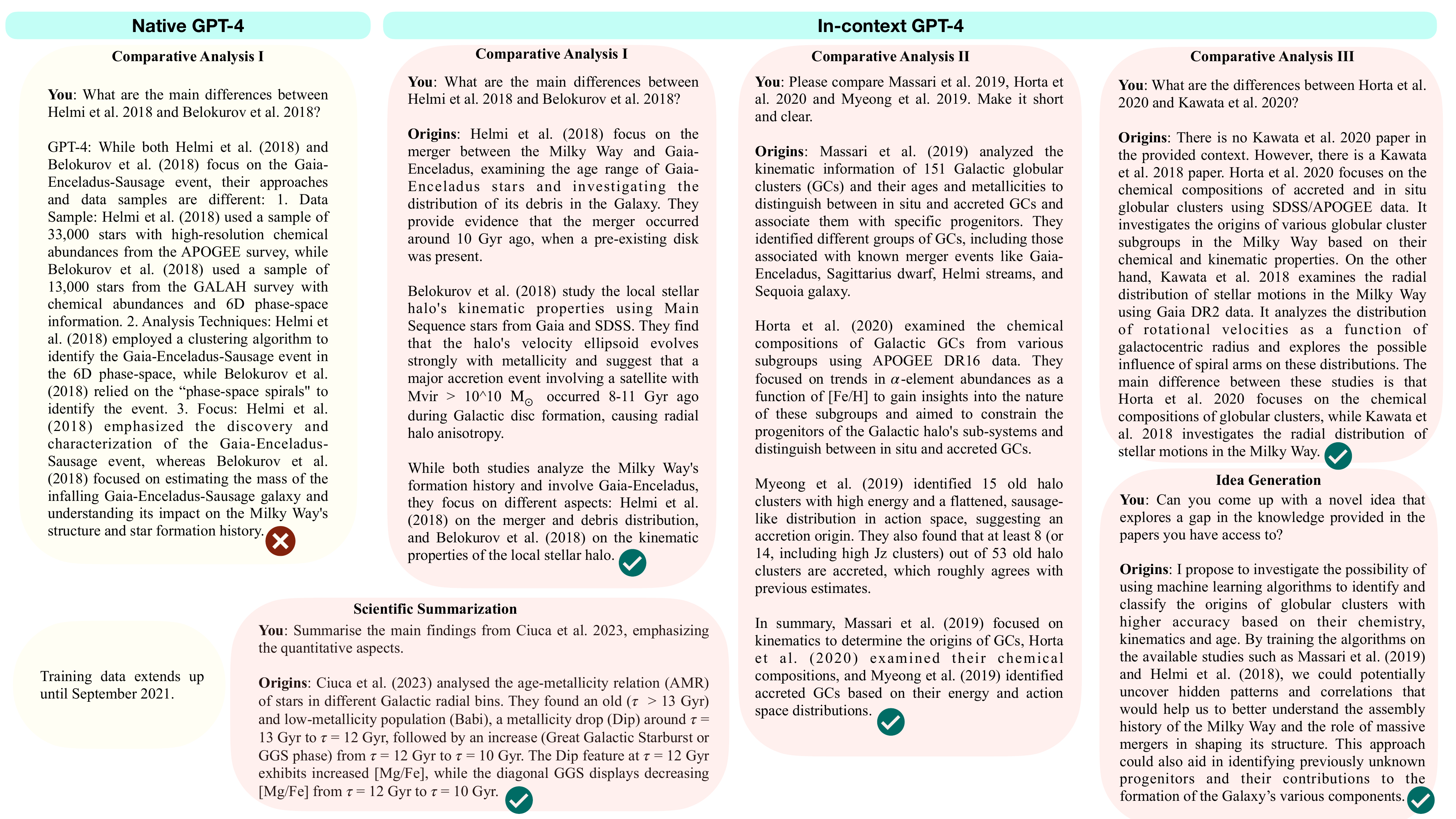}
 \caption{The GPT-4 model responses to expert-level questioning in no-context vs. multi-document context settings. No-context responses may lack recent findings and sometimes hallucinate details. For example, \cite{Belokurov_2018} did not use GALAH data. The in-context GPT-4 instance performs well in summarisation, study comparisons, and idea generation.  For example, the model accurately identifies that it only has access to \cite{Kawata_2018}, as shown in Comparative Analysis III.
\label{fig:drchattie}}
\end{figure}

We select ten papers in Galactic Archaeology, including \cite{Helmi_2018, Belokurov_2018, Kawata_2018, Massari2019, Myeong2019, Horta2020, Ciuca2023}. We extract the text from the PDF and divide it, then pass to the GPT-4 powered ChatGPT \citep{OpenAIGPT4}
using a \textbf{distillation prompt} that contains: `Distill each paragraph of the given text, maintaining the same number of paragraphs and structure. Limit the word count to 50\% of the original, and ensure references are included.' We essentially `compress' the paper to half its size. Unlike summarisation, this process ensures that the distilled paper inherits the original paper's paragraph structure and associated information. The distillation process allows us to use more processed papers efficiently for a context window with a given $\sim$8K token limit.

Our approach employs the \texttt{langchain} \citep{chase2022langchain} framework. The distilled input is embedded using OpenAI's \texttt{text-embedding-ada-002} model and stored in a large vector store. The expert query and chat history are processed by GPT-4 to generate a standalone question, which is also embedded. We 
then use FAISS (Facebook AI Similarity Search) \citep{johnson2019billion} to perform a similarity search between the embedded question and the input and retrieve a relevant context for the model. Finally, the GPT-4 model, accessed through the OpenAI API, uses this context and the standalone question to generate a response. The model uses the following \textbf{system prompt}: `Engage in insightful conversations with humans, providing meaningful, concise answers based on the provided documents. Include pertinent study citations, such as Example et al. (2020).'
\section{Results and conclusion}
\label{results}

We explore the performance of GPT-4 when responding to expert-level inquiries concerning summarization, comparative analysis, and idea generation, as illustrated in Figure \ref{fig:drchattie}. The native model may generate imprecise information through a phenomenon called `hallucination' \citep[e.g.,][]{shuster2021retrieval,ji2022survey, Peng}. However, when given access to context, the model's performance significantly improves, as shown by the different responses in the first two left panels of Figure \ref{fig:drchattie} between the native and in-context GPT-4, which we denote by Origins\footnote{We provide an example of a conversation with Origins, the in-galactic-archaeology-context GPT-4 model, at \url{https://www.youtube.com/watch?v=cufBNDDBgJ4}.}. The in-context prompting allows the model to explore connections and differences across multiple papers, as exemplified by the Comparative Analysis II in Figure \ref{fig:drchattie}, where the model correctly identifies the difference in the focus of the two papers provided.

The Comparative Analysis III in Figure \ref{fig:drchattie} demonstrates that the in-context GPT-4 model can identify if it has access to a particular paper, indexed by the first author(s) and year of publication, and then uses that to answer the question. We recognize the potential for idea generation as shown in the lower right panel, with the model identifying a possible new link between machine learning, which was employed to study the Milky Way disc by \cite{Ciuca2023}, with the vast available data for globular clusters.

To conclude, this research note emphasizes the importance of utilizing in-context prompting with large language models to engage with Astronomy papers effectively. In the future, we plan to investigate the model's response quality as a function of the number of input papers and other variables, such as the focus of the papers and the broadness of the research. On the technical aspect, we plan to explore how the responses vary with the distillation level and how they compare to those from a fine-tuned model.

\begin{acknowledgments}
The authors thank the OpenAI Team for access to the GPT-4 model. IC is grateful for the Joint Jubilee Fellowship at the ANU. YST acknowledges financial support from the Australian Research Council through DECRA Fellowship DE220101520. IC is grateful to Lachezar Todorov, who provided access to the GPT-4 API and to the \texttt{langchain} community, in particular Hamel Husain and Mayo Oshin, for their open-source code. IC is grateful to Dr Josh Peek and Dr Kartheik Iyer for their insightful comments. IC and YST  thank the Kavli Institute for Theoretical Physics, where this research was undertaken as part of the `Building a Physical Understanding of Galaxy Evolution with Data-driven Astronomy'. This research was in part supported by the National Science Foundation under Grant No. NSF PHY-1748958.
\end{acknowledgments}

%

\vspace{5mm}


\software{OpenAI GPT-4 API \citep{OpenAIGPT4},
           \texttt{langchain} \citep{chase2022langchain},
           FAISS  \citep{johnson2019billion}.
          }

\bibliography{ms}{}

\begin{thebibliography}{}
\expandafter\ifx\csname natexlab\endcsname\relax\def\natexlab#1{#1}\fi
\providecommand{\url}[1]{\href{#1}{#1}}
\providecommand{\dodoi}[1]{doi:~\href{http://doi.org/#1}{\nolinkurl{#1}}}
\providecommand{\doeprint}[1]{\href{http://ascl.net/#1}{\nolinkurl{http://ascl.net/#1}}}
\providecommand{\doarXiv}[1]{\href{https://arxiv.org/abs/#1}{\nolinkurl{https://arxiv.org/abs/#1}}}

\bibitem[{{Belokurov} {et~al.}(2018){Belokurov}, {Erkal}, {Evans}, {Koposov},
  \& {Deason}}]{Belokurov_2018}
{Belokurov}, V., {Erkal}, D., {Evans}, N.~W., {Koposov}, S.~E., \& {Deason},
  A.~J. 2018, \mnras, 478, 611, \dodoi{10.1093/mnras/sty982}

\bibitem[{Brown {et~al.}(2020)Brown, Mann, Ryder, Subbiah, Kaplan, Dhariwal,
  Neelakantan, Shyam, Sastry, Askell, Agarwal, Herbert-Voss, Krueger, Henighan,
  Child, Ramesh, Ziegler, Wu, Winter, Hesse, Chen, Sigler, Litwin, Gray, Chess,
  Clark, Berner, McCandlish, Radford, Sutskever, \& Amodei}]{brown2020language}
Brown, T.~B., Mann, B., Ryder, N., {et~al.} 2020, Language Models are Few-Shot
  Learners.
\newblock \doarXiv{2005.14165}

\bibitem[{Bubeck {et~al.}(2023)Bubeck, Chandrasekaran, Eldan, Gehrke, Horvitz,
  Kamar, Lee, Lee, Li, Lundberg, Nori, Palangi, Ribeiro, \&
  Zhang}]{bubeck2023sparks}
Bubeck, S., Chandrasekaran, V., Eldan, R., {et~al.} 2023, Sparks of Artificial
  General Intelligence: Early experiments with GPT-4.
\newblock \doarXiv{2303.12712}

\bibitem[{Chase(2022)}]{chase2022langchain}
Chase, H. 2022, LangChain.
\newblock \url{https://github.com/hwchase17/langchain}

\bibitem[{{Chowdhery} {et~al.}(2022){Chowdhery}, {Narang}, {Devlin}, {Bosma},
  {Mishra}, {Roberts}, {Barham}, {Chung}, {Sutton}, {Gehrmann}, {Schuh}, {Shi},
  {Tsvyashchenko}, {Maynez}, {Rao}, {Barnes}, {Tay}, {Shazeer}, {Prabhakaran},
  {Reif}, {Du}, {Hutchinson}, {Pope}, {Bradbury}, {Austin}, {Isard}, {Gur-Ari},
  {Yin}, {Duke}, {Levskaya}, {Ghemawat}, {Dev}, {Michalewski}, {Garcia},
  {Misra}, {Robinson}, {Fedus}, {Zhou}, {Ippolito}, {Luan}, {Lim}, {Zoph},
  {Spiridonov}, {Sepassi}, {Dohan}, {Agrawal}, {Omernick}, {Dai},
  {Sankaranarayana Pillai}, {Pellat}, {Lewkowycz}, {Moreira}, {Child},
  {Polozov}, {Lee}, {Zhou}, {Wang}, {Saeta}, {Diaz}, {Firat}, {Catasta}, {Wei},
  {Meier-Hellstern}, {Eck}, {Dean}, {Petrov}, \& {Fiedel}}]{Chowdhery2022}
{Chowdhery}, A., {Narang}, S., {Devlin}, J., {et~al.} 2022, arXiv e-prints,
  arXiv:2204.02311, \dodoi{10.48550/arXiv.2204.02311}

\bibitem[{{Ciuc{\u{a}}} {et~al.}(2023){Ciuc{\u{a}}}, {Kawata}, {Ting}, {Grand},
  {Miglio}, {Hayden}, {Baba}, {Fragkoudi}, {Monty}, {Buder}, \&
  {Freeman}}]{Ciuca2023}
{Ciuc{\u{a}}}, I., {Kawata}, D., {Ting}, Y.-S., {et~al.} 2023, \mnras,
  \dodoi{10.1093/mnrasl/slad033}

\bibitem[{{Devlin} {et~al.}(2018){Devlin}, {Chang}, {Lee}, \&
  {Toutanova}}]{Devlin2018}
{Devlin}, J., {Chang}, M.-W., {Lee}, K., \& {Toutanova}, K. 2018, arXiv
  e-prints, arXiv:1810.04805, \dodoi{10.48550/arXiv.1810.04805}

\bibitem[{Helmi {et~al.}(2018)Helmi, Babusiaux, Koppelman, Massari, Veljanoski,
  \& Brown}]{Helmi_2018}
Helmi, A., Babusiaux, C., Koppelman, H.~H., {et~al.} 2018, Nature, 563,
  85–88, \dodoi{10.1038/s41586-018-0625-x}

\bibitem[{{Horta} {et~al.}(2020){Horta}, {Schiavon}, {Mackereth}, {Beers},
  {Fern{\'a}ndez-Trincado}, {Frinchaboy}, {Garc{\'\i}a-Hern{\'a}ndez},
  {Geisler}, {Hasselquist}, {J{\"o}nsson}, {Lane}, {Majewski},
  {M{\'e}sz{\'a}ros}, {Bidin}, {Nataf}, {Roman-Lopes}, {Nitschelm},
  {Vargas-Gonz{\'a}lez}, \& {Zasowski}}]{Horta2020}
{Horta}, D., {Schiavon}, R.~P., {Mackereth}, J.~T., {et~al.} 2020, \mnras, 493,
  3363, \dodoi{10.1093/mnras/staa478}

\bibitem[{{Howard} \& {Ruder}(2018)}]{howard2018}
{Howard}, J., \& {Ruder}, S. 2018, arXiv e-prints, arXiv:1801.06146,
  \dodoi{10.48550/arXiv.1801.06146}

\bibitem[{{Hu} {et~al.}(2021){Hu}, {Shen}, {Wallis}, {Allen-Zhu}, {Li}, {Wang},
  {Wang}, \& {Chen}}]{Hu2021}
{Hu}, E.~J., {Shen}, Y., {Wallis}, P., {et~al.} 2021, arXiv e-prints,
  arXiv:2106.09685, \dodoi{10.48550/arXiv.2106.09685}

\bibitem[{Ji {et~al.}(2022)Ji, Lee, Frieske, Yu, Su, Xu, Ishii, Bang, Dai,
  Madotto, \& Fung}]{ji2022survey}
Ji, Z., Lee, N., Frieske, R., {et~al.} 2022, Survey of Hallucination in Natural
  Language Generation.
\newblock \doarXiv{2202.03629}

\bibitem[{Johnson {et~al.}(2019)Johnson, Douze, \&
  J{\'e}gou}]{johnson2019billion}
Johnson, J., Douze, M., \& J{\'e}gou, H. 2019, IEEE Transactions on Big Data,
  7, 535

\bibitem[{{Kawata} {et~al.}(2018){Kawata}, {Baba}, {Ciuc{\v{a}}}, {Cropper},
  {Grand}, {Hunt}, \& {Seabroke}}]{Kawata_2018}
{Kawata}, D., {Baba}, J., {Ciuc{\v{a}}}, I., {et~al.} 2018, \mnras, 479, L108,
  \dodoi{10.1093/mnrasl/sly107}

\bibitem[{{Massari} {et~al.}(2019){Massari}, {Koppelman}, \&
  {Helmi}}]{Massari2019}
{Massari}, D., {Koppelman}, H.~H., \& {Helmi}, A. 2019, \aap, 630, L4,
  \dodoi{10.1051/0004-6361/201936135}

\bibitem[{{Myeong} {et~al.}(2018){Myeong}, {Evans}, {Belokurov}, {Sanders}, \&
  {Koposov}}]{Myeong2019}
{Myeong}, G.~C., {Evans}, N.~W., {Belokurov}, V., {Sanders}, J.~L., \&
  {Koposov}, S.~E. 2018, \apjl, 863, L28, \dodoi{10.3847/2041-8213/aad7f7}

\bibitem[{OpenAI(2023)}]{OpenAIGPT4}
OpenAI. 2023, ArXiv, abs/2303.08774

\bibitem[{{Peng} {et~al.}(2023){Peng}, {Galley}, {He}, {Cheng}, {Xie}, {Hu},
  {Huang}, {Liden}, {Yu}, {Chen}, \& {Gao}}]{Peng}
{Peng}, B., {Galley}, M., {He}, P., {et~al.} 2023, arXiv e-prints,
  arXiv:2302.12813, \dodoi{10.48550/arXiv.2302.12813}

\bibitem[{Pfeiffer {et~al.}(2020)Pfeiffer, Vuli{\'c}, Gurevych, \&
  Ruder}]{pfeiffer2020}
Pfeiffer, J., Vuli{\'c}, I., Gurevych, I., \& Ruder, S. 2020, in Proceedings of
  the 2020 Conference on Empirical Methods in Natural Language Processing
  (EMNLP) (Online: Association for Computational Linguistics), 7654--7673,
  \dodoi{10.18653/v1/2020.emnlp-main.617}

\bibitem[{Shuster {et~al.}(2021)Shuster, Poff, Chen, Kiela, \&
  Weston}]{shuster2021retrieval}
Shuster, K., Poff, S., Chen, M., Kiela, D., \& Weston, J. 2021, Retrieval
  Augmentation Reduces Hallucination in Conversation.
\newblock \doarXiv{2104.07567}

\bibitem[{Wei {et~al.}(2022)Wei, Bosma, Zhao, Guu, Yu, Lester, Du, Dai, \&
  Le}]{wei2022finetuned}
Wei, J., Bosma, M., Zhao, V.~Y., {et~al.} 2022, Finetuned Language Models Are
  Zero-Shot Learners.
\newblock \doarXiv{2109.01652}

\bibitem[{White(2023)}]{white2023}
White, A. 2023, Paper QA.
\newblock \url{https://github.com/whitead/paper-qa}

\end{thebibliography}
\bibliographystyle{aasjournal}



\end{document}